# An Immune Inspired Network Intrusion Detection System Utilising Correlation Context

Gianni Tedesco[*]  Uwe Aickelin[*]

[*]School of Computer Science & IT (ASAP)
University of Nottingham
NG8 1BB
gxt@cs.nott.ac.uk

**Abstract**

Network Intrusion Detection Systems (NIDS) are computer systems which monitor a network with the aim of discerning malicious from benign activity on that network. While a wide range of approaches have met varying levels of success, most IDSs rely on having access to a database of known attack signatures which are written by security experts. Nowadays, in order to solve problems with false positive alerts, correlation algorithms are used to add additional structure to sequences of IDS alerts. However, such techniques are of no help in discovering novel attacks or variations of known attacks, something the human immune system (HIS) is capable of doing in its own specialised domain. This paper presents a novel immune algorithm for application to the IDS problem. The goal is to discover packets containing novel variations of attacks covered by an existing signature base.

## 1 Introduction

Network intrusion detection systems (NIDS) are usually based on a fairly low level model of network traffic. While this is good for performance it tends to produce results which make sense on a similarly low level which means that a fairly sophisticated knowledge of both networking technology and infiltration techniques is required to understand them.

Intrusion alert correlation systems attempt to solve this problem by post-processing the alert stream from one or many intrusion detection sensors (perhaps even heterogeneous ones). The aim is to augment the somewhat one-dimensional alert stream with additional structure. Such structural information clusters alerts in to scenarios sequences of low level alerts corresponding to a single logical threat.

A common model for intrusion alert correlation algorithms is that of the attack graph. Attack graphas are directed acyclic graphs (DAGs) that attempt to represent the various types of alerts in terms of their prerequisites and consequences. Typically an attack graph is created by an expert from a priori information about attacks. The attack graph enables a correlation component to link a given alert with a previous alert by tracking back to find alerts whose consequences imply the current alerts prerequisites. Another feature is that if the correlation algorithm is run in reverse, predictions of future attacks can be obtained.

In implementing basic correlation algorithms using attack graphs, it was discovered that the output could be poor when the underlying IDS produced false negative alerts. This could cause scenarios to be split apart as evidence suggestive of a link between two scenarios is missing. This problem has been addressed in various systems by adding the ability to hypothesise the existence of the missing alerts in certain cases. Ning et al (2004) go as far as to use out of band data from a raw audit log of network traffic to help confirm or deny such hypotheses.

While the meaning of correlated alerts and predicted alerts is clear, hypothesisd results are less easy to interpret. Presence of hypothesised alerts could mean more than just losing an alert, it could mean either of:

1. The IDS missed the alert due to some noise, packet loss, or other low level sensor problem

2. The IDS missed the alert because a novel variation of a known attack was used

3. The IDS missed the alert, because something not covered by the attack graph happened (totally new exploit, or new combination of known exploits)

This work is motivated specifically by the problem of finding novel variations of attacks. In our case a variation is determined to be an attack which exploits the same vector as an attack detected by an existing rule. The basic approach is to apply AIS techniques to detect packets which contain such variations. A correlation algorithm is taken advantage of to provide additional safe/dangerous context signals to the AIS which would enable it to decide which packets to examine. The work aims to integrate a novel AIS component with existing intrusion detection and alert correlation systems in order to gain additional detection capability.

## 2 Intrusion Alert Correlation

Although the exact implementation details of attack graphs algorithms vary, the basic correlation algorithm takes an alert and an output graph, and modifies the graph by addition of vertices and/or edges to produce an updated output graph reflecting the current state of the monitored network system.

For the purposes of discussion, an idealised form of correlation output will be defined which hides specific details of the correlation algorithm from the AIS component. This model, while fairly simple, adequately maps to current state of the art correlation algorithms. Due to space constraints we do not describe the full model here.

## 3 Danger Theory

The advent of Polly Matzingers Danger theory in has inspired a great deal of research in to the functioning of the innate immune system. A subsystem of the human immune system (HIS) which is apparently able to distinguish between benign and pathogenic material within the organism and initiate an adaptive immune response.

For this purpose our "libtissue" AIS framework, a product of a danger theory project (Aickelin et al, 2003), will model a number of innate immune system components such as dendritic cells in order to direct an adaptive T-Cell based response.

Dendritic cells (henceforth DCs) are of a class of cells in the immune system known as antigen presenting cells. They differ from other cells in this class in that this is their sole discernable function. As well as being able to absorb and present antigenic material DCs are also well adapted to detecting a set of endogenous and exogenous signals. These biological signals are abstracted in our system under the following designations:

1. Safe: Indicates a safe context for developing toleration

2. Danger: Indicates a change in behaviour that could be considered pathological

3. Pathogen Associated Molecular Pattern (PAMP): Known to be dangerous

All of these environmental circumstances, or inputs, are factors in the life cycle of the DC. A sufficient concentration of signals may trigger maturation along one of two differented pathways. One of which is associated with a reactive and the other with a tolerogenic T-cell response.

In the proposed system, DCs are seen as living among the IDS environment. This is achieved by wiring up their environmental inputs to certain changes in the IDS output state. Populations of DCs are tied to the prediction vertices in the correlation graph, one DC for each predicted attack. Packets matching the prediction criteria of such a vertex are injested by the corresponding DC.

A prediction veretex can either be upgraded to an exploit vertex, changed to a hypothesised vertex, or be deleted depending on subsequent alerts. These possibilities will result in either a PAMP, danger or safe signal respectively.

These signals initiate maturation and consequent migration of the DC to a virtual lymph node where they are exposed to a population of T-cells generated using the IDSs signature base in much the same way as in a gene library. This is combined with partial matching algorithms to find a T-cells to bind to the antigen being presented by the DC.

Upon successful binding, the original packet corresponding to the culprit antigen is tagged and logged much like a normal alert.